\documentclass{article}

\usepackage{amsmath}
\usepackage{amsfonts}
\usepackage{amsthm}
\usepackage{amssymb}
\usepackage[space]{cite}
\usepackage{fullpage}
\usepackage{graphicx}
\usepackage{hyperref}
\usepackage{mathrsfs}

\DeclareMathOperator{\defeq}{\stackrel{\text{def\textsuperscript{\underline{n}}}}{=}}
\DeclareMathOperator{\Span}{\text{span}}

\begin{document}

\author{Andrzej S. Kucik\thanks{\url{andrzej.kucik@gmail.com}} \; and \, Konstantin Korovin\thanks{\url{korovin@cs.man.ac.uk}} \\ \\ \small{\textit{School of Computer Science, University of Manchester}}}
\title{Premise selection with neural networks and distributed representation of features}
\date{}
\maketitle

\begin{abstract}
We present the~problem of selecting relevant premises for a~proof of a~given statement. When stated as a~\emph{binary classification task} for pairs (conjecture, axiom), it can be efficiently solved using \emph{artificial neural networks}. The~key difference between our advance to solve this problem and previous approaches is the~use of just \emph{functional signatures} of premises. To further improve the~performance of the~model, we use \emph{dimensionality reduction} technique, to replace long and sparse signature vectors with their compact and dense embedded versions. These are obtained by firstly defining the~concept of a~\emph{context} for each functor symbol, and then training a~simple neural network to predict the~distribution of other functor symbols in the~context of this functor. After training the~network, the~output of its hidden layer is used to construct a~lower dimensional embedding of a~functional signature (for each premise) with \emph{distributed representation of features}. This allows us to use 512-dimensional embeddings for conjecture-axiom pairs, containing enough information about the~original statements to reach the~accuracy of 76.45\% in \emph{premise selection task}, only with simple two-layer densely connected neural networks. \\
\textbf{Keywords:} artificial neural networks, automated theorem proving, curse of dimensionality, deep learning, distributed representation of features, dimensionality reduction, machine learning, premise selection, vector embeddings. \\
\textbf{Mathematics Subject Classification (2010):} \href{https://mathscinet.ams.org/msc/msc2010.html?t=03B35&btn=Current}{03B35}, \href{https://mathscinet.ams.org/msc/msc2010.html?t=68T05&btn=Current}{68T05}, \href{https://mathscinet.ams.org/msc/msc2010.html?t=68T15&btn=Current}{68T15}.
\end{abstract}

\section{Introduction}
Automated theorem provers (ATPs) and proof assistants helped with formal verification of many well-known mathematical theorems, most notable examples being the~proofs of four colour theorem \cite{gonthier2008} and Kepler conjecture \cite{hales2017}. However, they face two significant limitations.

First of them is the~fact that they can only employ formalised mathematics. Most of the~corpora of mathematical results, although written in a~relatively formal manner, is still only available as natural language texts, and there exist no efficient semantic or formal parsers to translate them into the~machine language. This makes many important theorems (especially those published recently) unavailable for automated systems. Secondly, there is no formal method of emulating human intuition in choosing the~relevant, already known facts (i.e. \emph{premise selection}) and strategies for the~proof. This makes even simple and intuitive proofs intractable for ATPs.

Recent advancements in another field of artificial intelligence -- \emph{machine learning}, and in particular \emph{neural networks} (also known by their rebadged name: \emph{deep learning}) give us hope that this issues can be resolved. This is because they proved to be very successful in the~fields previously reserved almost exclusively for formal methods, such as strategy board games -- most notably the~game of GO \cite{alphago}.

The~first attempt to employ deep learning to automated theorem proving was made in \cite{alemi2016}, where the~neural network models are trained on pairs of first-order logic axioms and conjectures to determine which axiom is most likely to be relevant in constructing an~automated proof by the~prover. In this paper we build on the~foundation laid therein, showing that selecting an~appropriate representation of premises can greatly simplify the~problem, allowing us to use much simpler neural networks and, consequently, make the~decision in much shorter time.

In the~context of theorem proving, deep learning techniques were also recently used for example in \cite{allamanis2017, cai2017, cai2017', chojecki2017, chollet2017, deng2017, peng2017, whalen2016}.

The~code for neural network architectures presented in this paper, as well as for processing of the~input data, is available at \cite{kucik2018}.

\section{Datasets}

For this experiment we use a~dataset of 32,524 examples collected and organised by Josef Urban in \cite{jurban} for Mizar40 experiment and DeepMath experiment \cite{alemi2016}. Each example is of the~form:
\begin{verbatim}
     C tptpformula
     + tptpformula
          ...
     + tptpformula
     - tptpformula
          ...
     - tptpformula
\end{verbatim}
where {\tt tptpformula} is the~standard first-order logic TPTP representation, {\tt C} indicates a~conjecture, {\tt +}~a~premise that is needed for an~ATP proof of {\tt C}, and {\tt -} a~premise that is not required for the~proof but was highly ranked by a~$k$-nearest neighbour algorithm trained on the~proofs. For the~practical reasons dictated by the~theory of machine learning, there is roughly the~same number of useful and redundant facts associated with each conjecture.
In~total, we have 102,442 unique formulae across this dataset; 32,524 conjectures and 69,918 axioms. Every conjecture has 16 axioms assigned to it on average, with the~minimum being 2 and the~maximum -- 270.
We take each conjecture and corresponding axioms and form pairs (conjecture,~axiom), which will constitute our positive and negative examples (522,528 in total). In \cite{alemi2016} two alternative data representations were adopted; character-level and word-level representation. Both of these are problematic however. Premises have a~variable number of characters (5 to 84,299 with mean 304.5) and words/tokens (2 to 21,251, with mean 107.4) so they have to be either truncated or padded with zeros. The~character representation is given by an~80 dimensional one-hot encoding with a~maximum sequence length of 2048 characters, and the~word representation is obtained by word encoding of axiom embeddings computed by the~previously trained character-level model, and generating pseudo-random vectors (of the~same dimension) to encode tokens such as brackets and operators. The~maximum number of words is limited to 500. The~resulting datasets are sparse and highly dimensional, and some of the~important information is lost by restricting the~maximal number of words or characters. This obstructs the~performance of machine learning algorithms applied to it and, in case of artificial neural networks, imposes serious limitations on the~network architecture. In the~\hyperref[sec:distr]{next section} we present our approach to tackle this problem.

\section{Distributed representation of formulae signatures}
\label{sec:distr}

First, we limit the~information obtained from each formula to the~functor symbols, ignoring variable symbols (since they are essentially arbitrarily chosen characters), brackets, quantifiers, connectives, equality symbol etc. We will show that this information is sufficient to obtain accuracy similar to or higher than these obtained in the~earlier models. There is 13,217 unique functor symbols across all the~formulae in our dataset. Thus, to each of these functor symbols we can assign a~unique positive integer smaller or equal to 13,217 and then, for any formula $F$ in our dataset, we can represent its functional signature by a~13,217 dimensional vector, whose $i$\textsuperscript{th} coordinate is equal to the~number of occurrences of a~function $f_i$ in the~scope of $F$, associated with the~integer $i$. But this does not really solve the~problem present in previous approaches. Each formula usually contains only a~handful of functions, and hence, in this setting, it would be represented by a~sparse and long vector.

This phenomenon is very common, especially in natural language processing, and it is known as the~\emph{curse of dimensionality}. It can be solved by a~\emph{distributed representation of features}, and there are several algorithms which can efficiently create such representation, for example a~neural probabilistic language model from \cite {bengio2003} or t-SNE technique from \cite{maaten2008}. However, these methods are normally applied to textual (and hence temporal) data, and rely on the~concept of a~context, which is not defined for formulae signatures. We must therefore modify it to suit our setting.

Let $X$ be a~finite set of real, linearly independent vectors, and let $\mathcal{Y}$ be real vector spaces, which we will call \emph{input} and \emph{output}, respectively, and let $f~:~X~\rightarrow~\mathcal{Y}$. Suppose that we know the~values $f(x)$ for some (or for all) arguments $x \in X$ but we do not explicitly know what $f$ is. The~essence of machine learning is to determine $f$ or to find its approximation, and consequently, to find the~values $f(x)$ which were previously unknown. Usually $f$ cannot be (easily) represented algebraically, but we can find a~good approximation of $f$ as a~composition of simpler functions. This, in turn, is the~essence of neural network methods.

Suppose we have a~task $T_1$, to approximate a~function $f : X \rightarrow \mathcal{Y}$, and we do it by a~neural network $N$, which can be represented as a~composition of functions:
\begin{displaymath}
	\underbrace{X}_{\text{input set}} \quad \stackrel{L_1}{\longrightarrow} \quad L_1(X) \quad \stackrel{L_2}{\longrightarrow} \quad \cdots \quad \stackrel{L_n}{\longrightarrow} \quad (L_n \circ \cdots \circ L_1)(X) \;  \defeq \; N(X) \approx f(X) \quad \subset \quad \underbrace{\mathcal{Y}}_{\text{output space}},
\end{displaymath}
where $\approx$ denotes approximate equality with respect to some fixed cost function on $\mathcal{Y}$, and $L_1, \, \ldots, \, L_n$ are called \emph{hidden layers} (and they often are composite functions themselves).

If the~network $N$ performs well after some training, we may assume that the~first $k$ layers preserve and pass on some crucial information about the~input set $X$ to the~latter layers, needed to complete task $T_1$. Thus, we may fix the~parameters of $L_1, \, \ldots, L_k$ and only train those of $L_{k+1}, \, \ldots, L_n$, regarding $(L_k \circ \cdots \circ L_1)(X)$ as the~input set of a~new neural network $\tilde{N}$,
\begin{displaymath}
	\underbrace{(L_k \circ \cdots \circ L_1)(X)}_{\text{new input set}} \quad \stackrel{L_{k+1}}{\longrightarrow}\quad \cdots \quad \stackrel{L_n}{\longrightarrow} \quad (L_n \circ \cdots \circ L_1)(X) \;  \defeq \; \tilde{N}((L_k \circ \cdots \circ L_1)(X)) \approx f(X) \quad \subset \quad \underbrace{\mathcal{Y}}_{\text{output space}},
\end{displaymath}

Now, let $\tilde{\mathcal{X}} = \Span(X)$ and let $\tilde{\mathcal{Y}}$ be some (other) real vector space. Assume that we have a~new task $T_2$, to approximate a~function $g :~\tilde{\mathcal{X}}~\longrightarrow~\tilde{\mathcal{Y}}$. If we decide to solve task $T_2$ using neural networks, we need to remember that there is a~positive relationship between the~number of parameters of the~network and the~dimensionality of $\tilde{\mathcal{X}}$. If the~latter is big, then we must either choose a~simpler neural network architecture (potentially damaging its accuracy) or devote more time and hardware resources to the~training process, which is not always possible. In practice this is bypassed by dimensionality reducing data preprocessing, training only top layers of the~network in the~later phases of the~training or by loading pretrained layers (from some other tasks) and fixing them as the~initial layers for our network model, and only training the~layers on top of them. Pragmatic motivation of getting a~lower dimensional embedding for the~input space, as well as the~advantages of obtaining it either during the~main training process or beforehand - as a~separate learning task, is described in the~context of image classification and natural language processing, for example, in \cite{chollet2018}. 

Given that $X$ forms a~basis for $\tilde{\mathcal{X}}$, we may solve a~simplified version of task $T_2$ in the~following way. Every element $x$ in $\tilde{\mathcal{X}}$ can be represented as
\begin{displaymath}
	x = \sum_{i=1}^{|X|} a_i x_i,
\end{displaymath}
for some constants $a_1, \, \ldots, a_{|X|}~\in~\mathbb{R}$ and distinct vectors $x_1, \, \ldots, \, x_{|X|}~\in~X$. We use the~pretrained layers $L_1, \, \ldots, \, L_k$ to define
\begin{displaymath}
	\mathcal{L}(x) := \sum_{i=1}^{|X|} a_i (L_k \circ \cdots \circ L_1) (x_i),
\end{displaymath}
for all $1 \leq j \leq k$. Then we can approximate $g$ with a~neural network $M$, whose input space is given by $\mathcal{L}(\tilde{\mathcal{X}})$, subject to the~constraint
\begin{displaymath}
	g\left(\sum_{i=1}^{|X|} a_i x_i \right) \approx M\left(\sum_{i=1}^{|X|} a_i (L_n \circ \cdots \circ L_1)(x_i)\right), \qquad (\forall a_i \in \mathbb{R}, \, x_i \in X).
\end{displaymath}

Although we are still approximating $g$, if
\begin{displaymath}
	\dim((L_n \circ \cdots \circ L_1)(x)) < \dim(x) \qquad \qquad (\forall x \in X),
\end{displaymath}
then this embedding will reduce dimensionality of the input for a~neural network, allowing for a~more robust architecture of the~network $M$, as compared to networks using $\tilde{\mathcal{X}}$ as the~input. We can also experiment with several different network architectures, without having to obtain a~new, lower dimensional embedding each time.  And since $X$ is a basis for $\tilde{\mathcal{X}}$, training the layers $L_1, \, \ldots, \, L_k$ on it will be faster than training all $n$ layers ($n > k$) on a training set from $\tilde{\mathcal{X}}$ before freezing the first $k$ of them. That is, provided that the cardinality of $X$ is smaller than the cardinality of this training set.

In natural language processing we usually start with a~vocabulary (i.e. a~list of words) $V$. It can be represented as a~canonical basis $\{e_k\}_{k=1}^{|V|}$ for $\mathbb{R}^{|V|}$, that is the~set of $|V|$-dimensional vectors, with all entries equal to 0, but for the~$k$\textsuperscript{th} entry, where $k$ is the~index of a~word in $V$ which corresponds to $e_k$. Since such vocabularies are normally immensely large, before any language processing task, it is good to find a~lower dimensional, dense representation of $V$. It is usually done by extracting features from temporal context of each word. If we want to mimic the~same strategy for functional signatures of logical expressions, we must first define what a~context is in this setting, since a~functional signature, unlike a~sequence of words, is not a~temporal object.

First, note that if $P$ is a~premise, then we can represent its functional signature by
\begin{displaymath}
	\mathscr{S}(P) := \sum_{k = 1}^n \lambda_k(P) e_k,
\end{displaymath}
where $n$ is the~total number of unique function symbols across some corpus $\tilde{X}$ of premises' functional signatures, which contains $P$, $e_k$ is, again, the~unit vector corresponding to the~$k$\textsuperscript{th} function, and $\lambda_k(P)$ is the~number of occurrences of this function in the~scope of $P$. Now, let $f_i$ and $f_j$ be functions corresponding to $e_i$ and $e_j$ respectively, for $1 \leq i, \, j \leq n$. If there exists a~premise $Q \in X$ such that $\lambda_i(Q), \lambda_j(Q) \neq 0$, then we say that $f_j$ \emph{is in the~context of} $f_i$. We may represent the~frequency distribution $\varphi$ of functions in the~context of $f_i$ by
\begin{displaymath}
	\varphi(f_i) := \frac{\sum_{Q \in X \, : \, \lambda_i(Q) \neq 0} \mathscr{S}(Q)}{\left\| \sum_{Q \in X \, : \, \lambda_i(Q) \neq 0} \mathscr{S}(Q) \right\|_1}.
\end{displaymath}

We want to approximate $\varphi$ by a~neural network with two hidden layers:
\begin{displaymath}
	\sigma_2(W_2(\sigma_1(W_1 e_k + b_1)) + b_2) \approx \varphi(f_k) \qquad \qquad (\forall 1 \leq k \leq n),
\end{displaymath}
where $W_1$ and $W_2$ are $n' \times n$ and $n \times n'$ matrices, $b_1$ and $b_2$ are $n'$ and $n$ dimensional vectors, $\sigma_1$ and $\sigma_2$ are activation functions applied elementwise, and $n' < n$. After training this network, we may use this new, lower dimensional representation for the~functional signature of a~premise $P$:
\begin{displaymath}
	\tilde{\mathscr{S}}(P) := \sum_{k=1}^n \lambda_k(P) \sigma_1(W_1 e_k + b_1).
\end{displaymath}
In case when $\sigma_1$ is an~invertible function, we may even use:
\begin{equation}
	\tilde{\mathscr{S}}(P) := \sigma_1(W_1 \mathscr{S}(P)+ b_1). \tag{$*$}
	\label{eq:shrunksig}
\end{equation}
This is because approximating $f : X \rightarrow Y$ is equivalent to approximating $f \circ g^{-1} : g(X) \rightarrow Y$, whenever $g^{-1} : g(X) \rightarrow X$ is the~inverse of $g$.

In our experimental setting, $X$ is the~set of all one-hot encoded functor symbols ($|X|$ = 13,217), $n = 13,217$, $n' = 256$, $\sigma_1$ is the~hyperbolic tangent function, and $\sigma_2$ is the~softmax function. We use Keras library \cite{keras} to create neural network models for the~dimensionality reduced embeddings, as well as the~premise selection model in the~\hyperref[sec:prem_sel]{next section}. We initialise the~weight matrices using He uniform initialisation \cite{he2015}. We train the~network on the~13,217 dimensional identity matrix, taking batches of 4096 training examples, for 150 epochs, using RMSprop algorithm \cite{hinton2016} with decay of the~learning rate equal to $10^{-8}$. 

Usually, we train a~model on some set of training data, so that we can produce an~estimate of some unknown function, which can later be used to predict values of this function for data points which were previously unavailable. Here, we know the~contextual distribution for all the~functor symbols, and hence all the~values of $\varphi$, and we use the~network model to simply find a~less complex approximation of $\varphi$. For that reason we do not split the~data into training and validation tests (also, doing so would effectively exclude some parameters from training, as the~set $X$ is linearly independent). And since we want to approximate $\varphi$ as accurately as possible, given the~fixed number of network parameters, over-fitting is not discouraged. We do, however, shuffle the~data after every epoch, to allow for more distributed features in the~lower dimensional representation of our dataset. The~accuracy of this network, with respect to categorical crossentropy, reaches 84\% after the~training.  

Since the~set of all functional signatures $\tilde{X}$ contains only linear combinations of one-hot encoded representations of functor symbols (set $X$), and because $\tanh$ is an~invertible function, we can use \eqref{eq:shrunksig} as the~lower dimensional representation for functional signatures of premises in out dataset to develop a~premise selection model in the~\hyperref[sec:prem_sel]{next section}.

Alternatively, we could have used autoencoders (see \cite{liou2008, liou2014}), with the~same network architecture, to obtain a~lower dimensional representation of functional signatures directly. That is, we would want to find tensors $b_1, \, b_2, \, W_1, \, W_2$ (with the~same shapes as above) such that
\begin{displaymath}
	\mathscr{S}(P) \approx \sigma_2(W_2(\sigma_1(W_1 \mathscr{S}(P) + b_1)) + b_2),
\end{displaymath}
for all premises $P$ in our dataset. This na\"{i}ve approach saves us the~trouble of finding contextual distributions for all the~symbols, and normally it would be a~more natural choice of a~dimensionality reduction technique. However, empirically, the~premise selection models presented in the~\hyperref[sec:prem_sel]{next section} is less accurate if it uses this representation of data. Nonetheless, we included the~implementation of this alternative approach in \cite{kucik2018}, should an~interested reader wish to verify this.

\section{Premise selection model}
\label{sec:prem_sel}
From the~set of 32,524 conjectures and 69,918 axioms, we form a~set of 522,528 positive and negative examples, by concatenating the~new 256 dimensional signatures of corresponding axioms and premises. The~resulting set may be represented as $522,528 \times 512$ matrix (which is considerably smaller than $522,528 \times 26,434$ representation of full functional signatures, if the~dimensionality reduction had not been applied). From this tensor we randomly select 470,275 rows (90\%) for training, and use the~remaining 52,253 to form a~test set (10\%). We use the~standard regularisation of the~data, by computing the~mean and standard deviation along each column of the~training set, and subtracting this mean from each corresponding column, and dividing them by standard deviation.

We develop a~several variants of neural networks with two hidden densely connected layers. The~first layer has 64, 128, 256, 512 or 1024 output units, and the~number of the~output units of the~second layer lies in the~same range, provided that it is never bigger than the~number of output units of the~first layer. The~activation functions for these layers are the~rectified linear units (ReLU). Both of these hidden layers are followed by a~dropout layer \cite{hinton2014}, with the~dropout rate 0.5 - to reduce the~overfitting of the~model. The~output layer activation function is the~logistic sigmoid function, returning the~predicted probability that the~tested premise is relevant for some proof of the~tested conjecture. During the~development stage, we also extract 10\% of the~training data for validation of the~models. The~models are trained for up to 1500 epochs, on batches of 4096 examples using the~Adam optimiser \cite{kingma2015} (with the~learning rate $10^{-4}$) with respect to the~logistic loss function. The~training data is shuffled after each epoch. The~test results are presented in the~table below.

\begin{center}
\begin{tabular}{c | l  c c c c c}
& & & & \textbf{Layer 1} \\
\textbf{Layer 2} & & 64 & 128 & 256 & 512 & 1024 \\ \cline{2-7}
& loss & 0.5418 & 0.5295 & 0.5173 & 0.5292 & 0.5687 \\
64 & accuracy & 72.21\% & 72.75\% & 73.52\% & 74.57\% & 75.19\% \\ 
& \# of param.  & 37,057 & 73,985 & 147,841 & 295,553 & 590,977 \\
& & & \\
& loss & & 0.5315 & 0.5158 & 0.5224 & 0.5523 \\
128 & accuracy & & 72.94\% & 73.73\% & 74.49\% & 75.47\% \\
& \# of param.& & 82,305  & 164,535 & 328,449 & 656,641 \\
& & & \\
& loss & & & 0.5195 & 0.5135 &  0.5347 \\
256 & accuracy & & & 73.63\% & 74.19\% & 75.32\% \\
& \# of param.& & & 197,377 & 394,241 & 787,969 \\
& & & \\
& loss & &  & & 0.5095 & 0.5166\\
512 & accuracy & &  & & 74.58\% & 75.34\% \\
& \# of param.& & & & 525,825 & 1,050,625 \\
& & & \\
& loss & & & &  & 0.5024\\
1024 & accuracy & & &  & & 75.76\%\\
& \# of param.& & & & & 1,575,937 \\
\end{tabular}
\end{center}

As we can see above the~$64 \, \times \, 64$ (64 output units for each of the~hidden layers) reaches the~lowest accuracy out of these fourteen models. But is does so with comparatively few trainable parameters, which means that it can be trained in a~short time and it makes predictions quickly. The~$1024 \, \times \, 1024$ model has the~highest accuracy, but it requires the~biggest amount of parameters, so naturally it is significantly slower. Furthermore, it is also more prone to overfitting, which can be seen of the~graphs below.

\begin{center}
\includegraphics[width=.5\linewidth]{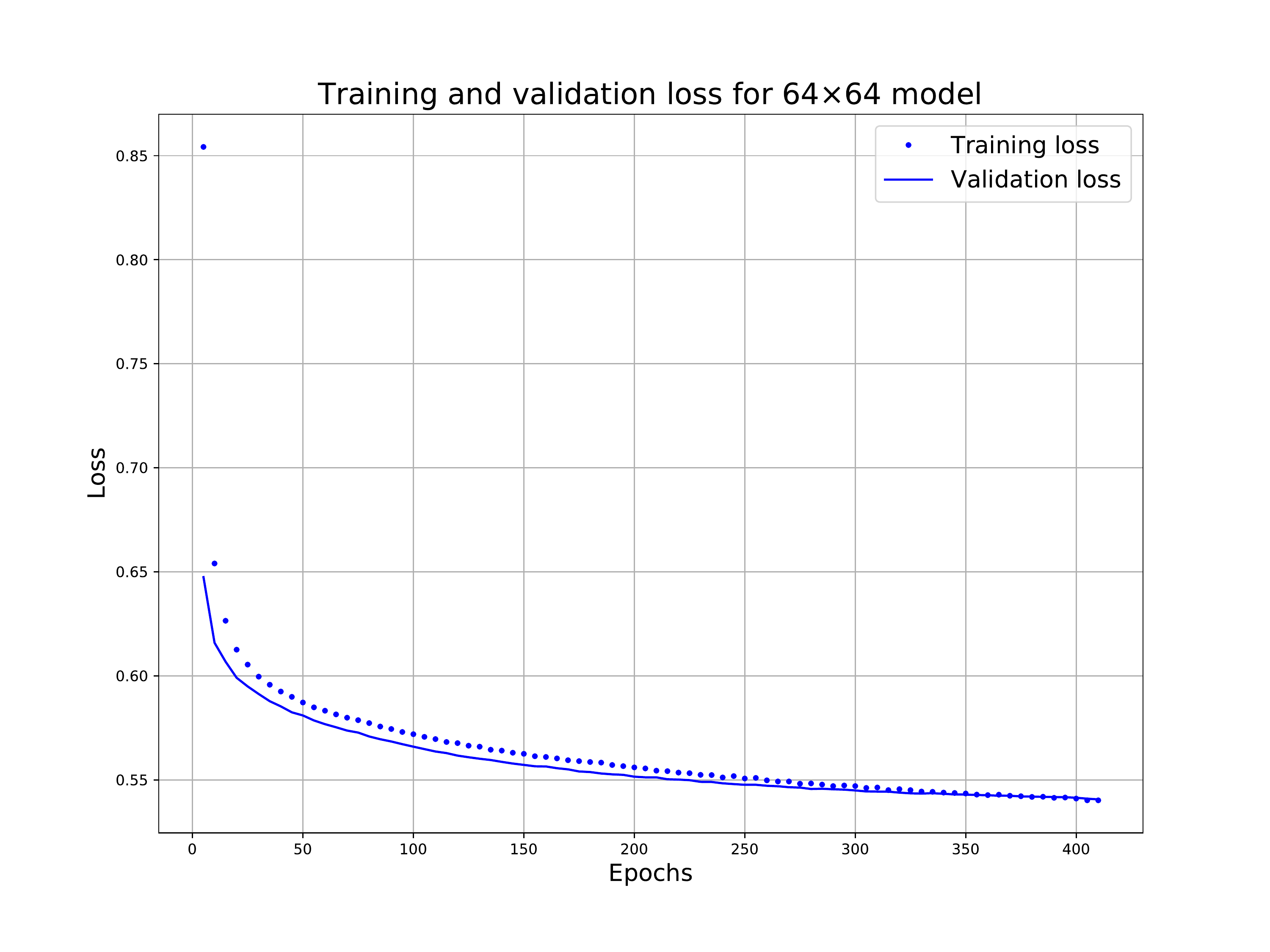}\includegraphics[width=.5\linewidth]{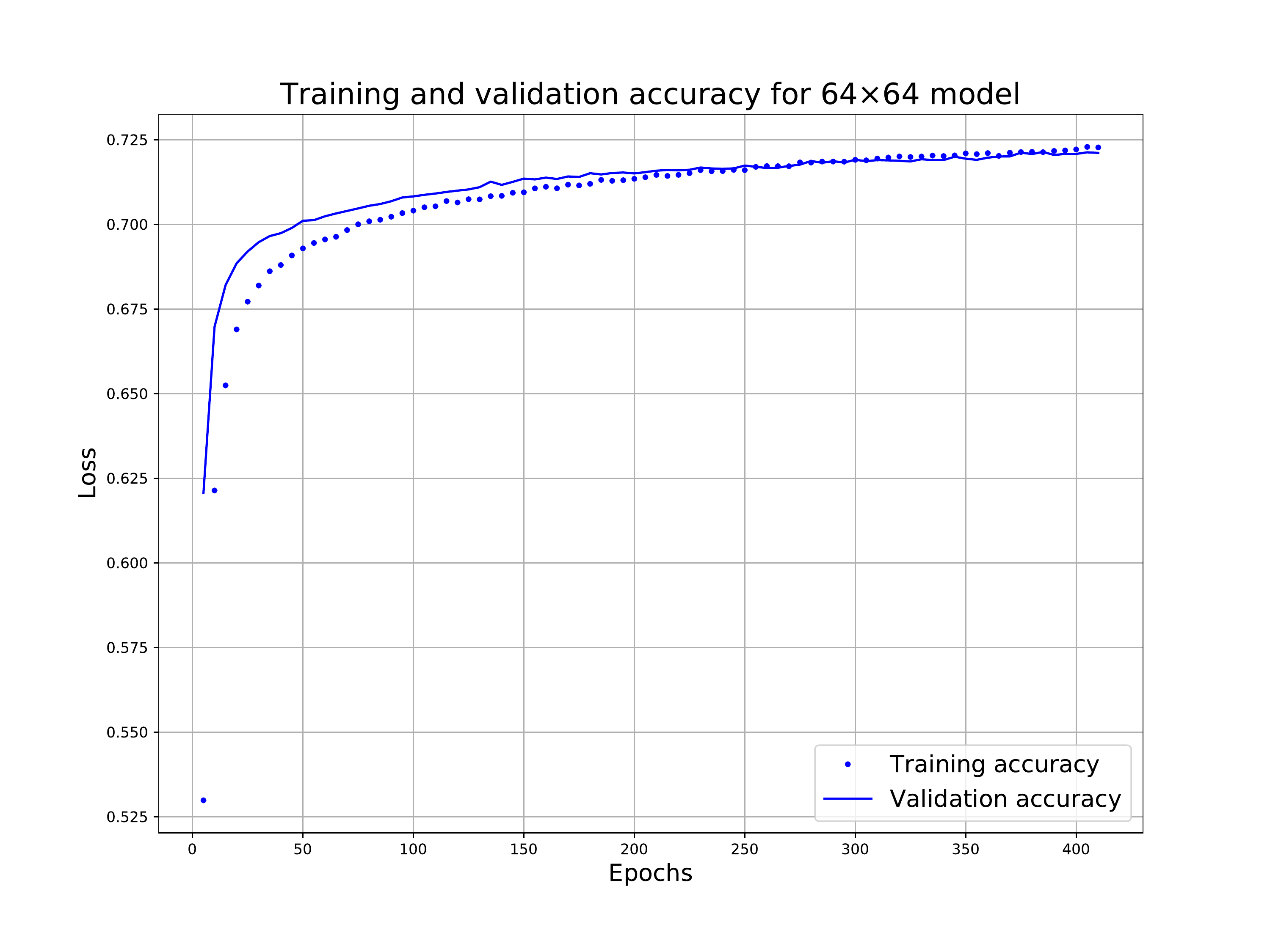} \\
\includegraphics[width=.5\linewidth]{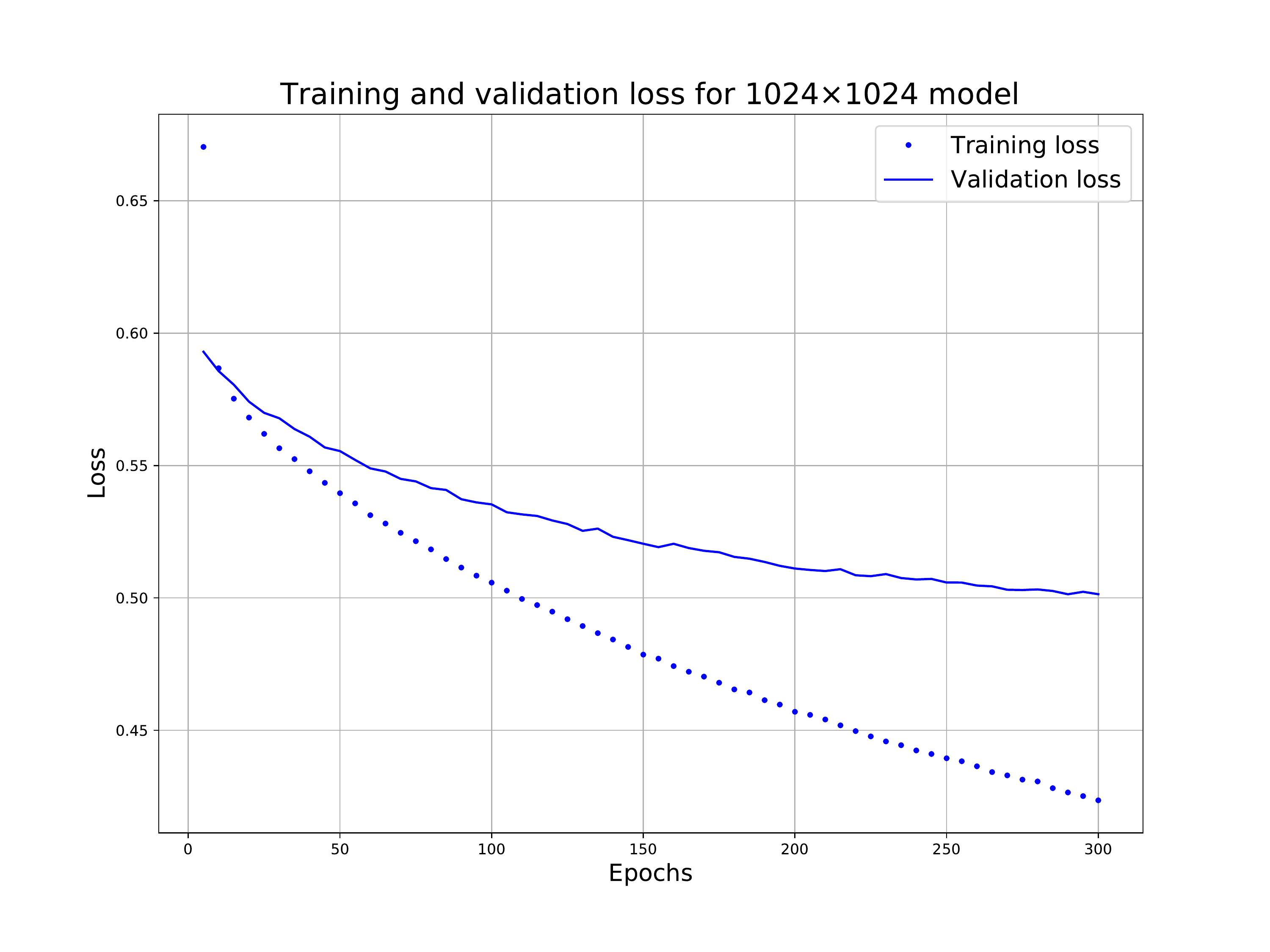}\includegraphics[width=.5\linewidth]{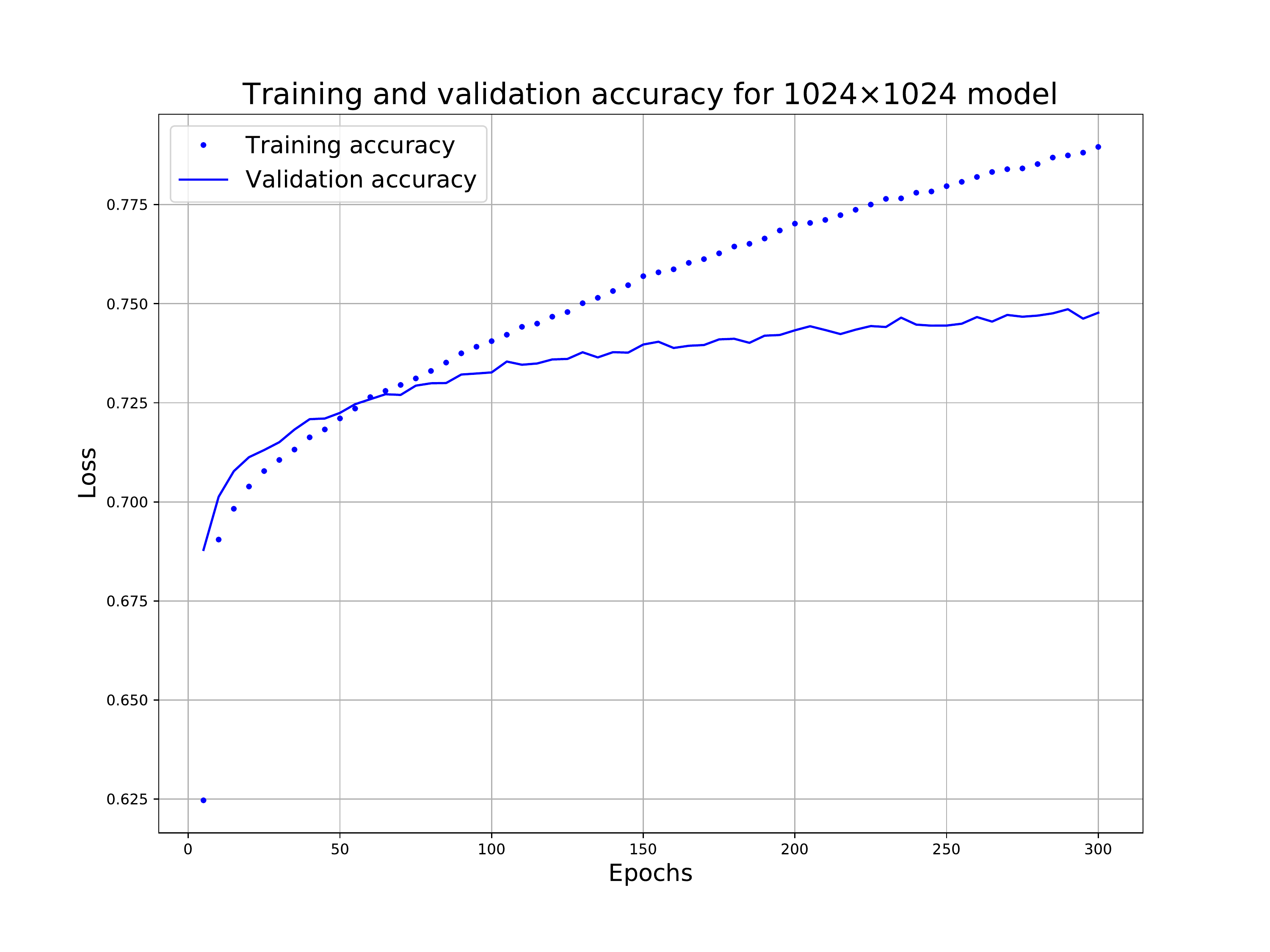}
\end{center}

Given the~trade-off between the~number of parameters (and hence the~computation time) and the~accuracy of the~model, one should choose the~most suitable model carefully. We chose $64 \, \times \, 64, \; 256 \, \times \, 256, \; 512 \, \times \, 128$ and $1024 \, \times \, 1024$, and trained them again, this time for 2500 epochs and without extracting any validation data. The~results are presented below.

\begin{center}
\includegraphics[width=.5\linewidth]{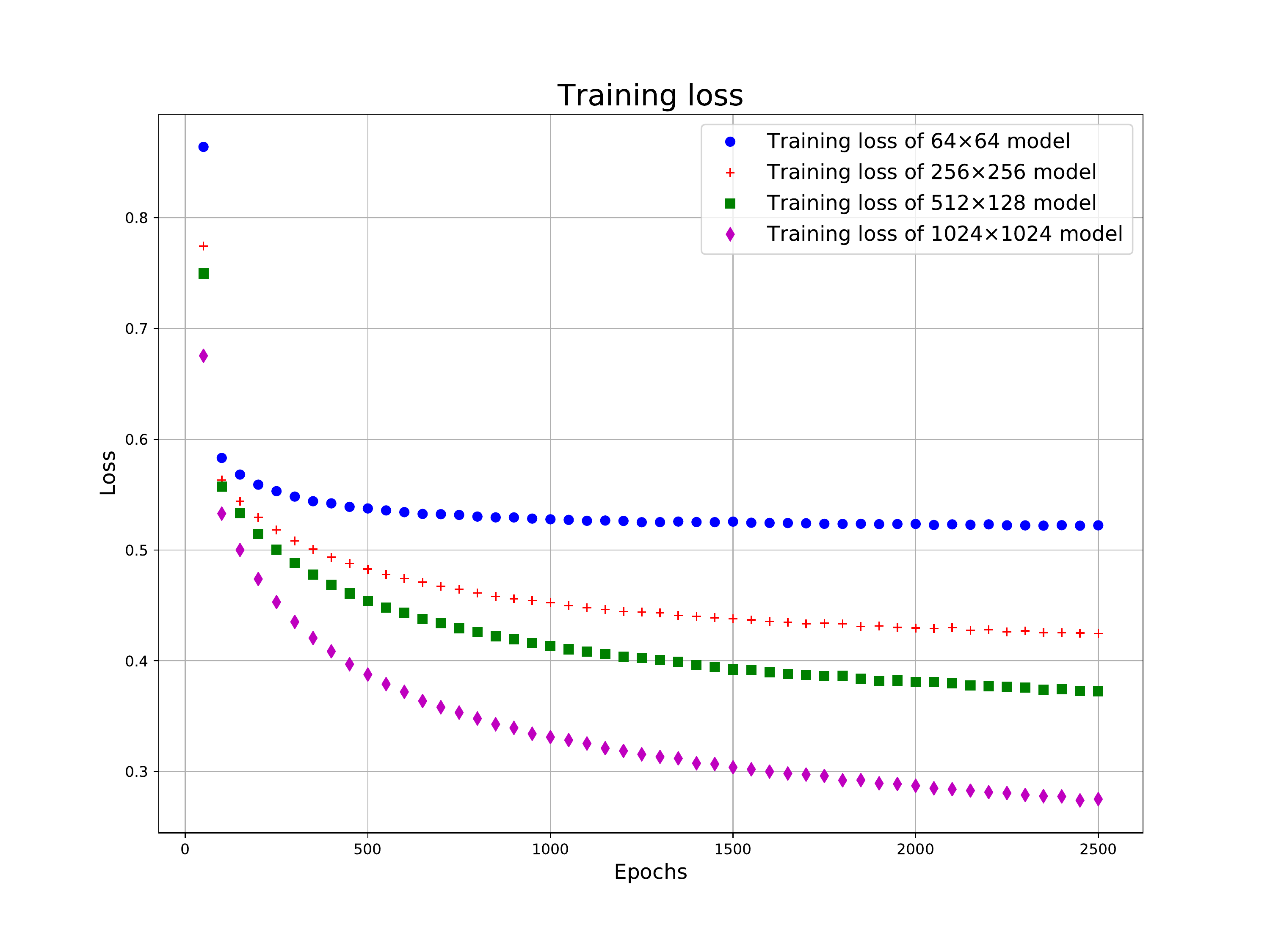}\includegraphics[width=.5\linewidth]{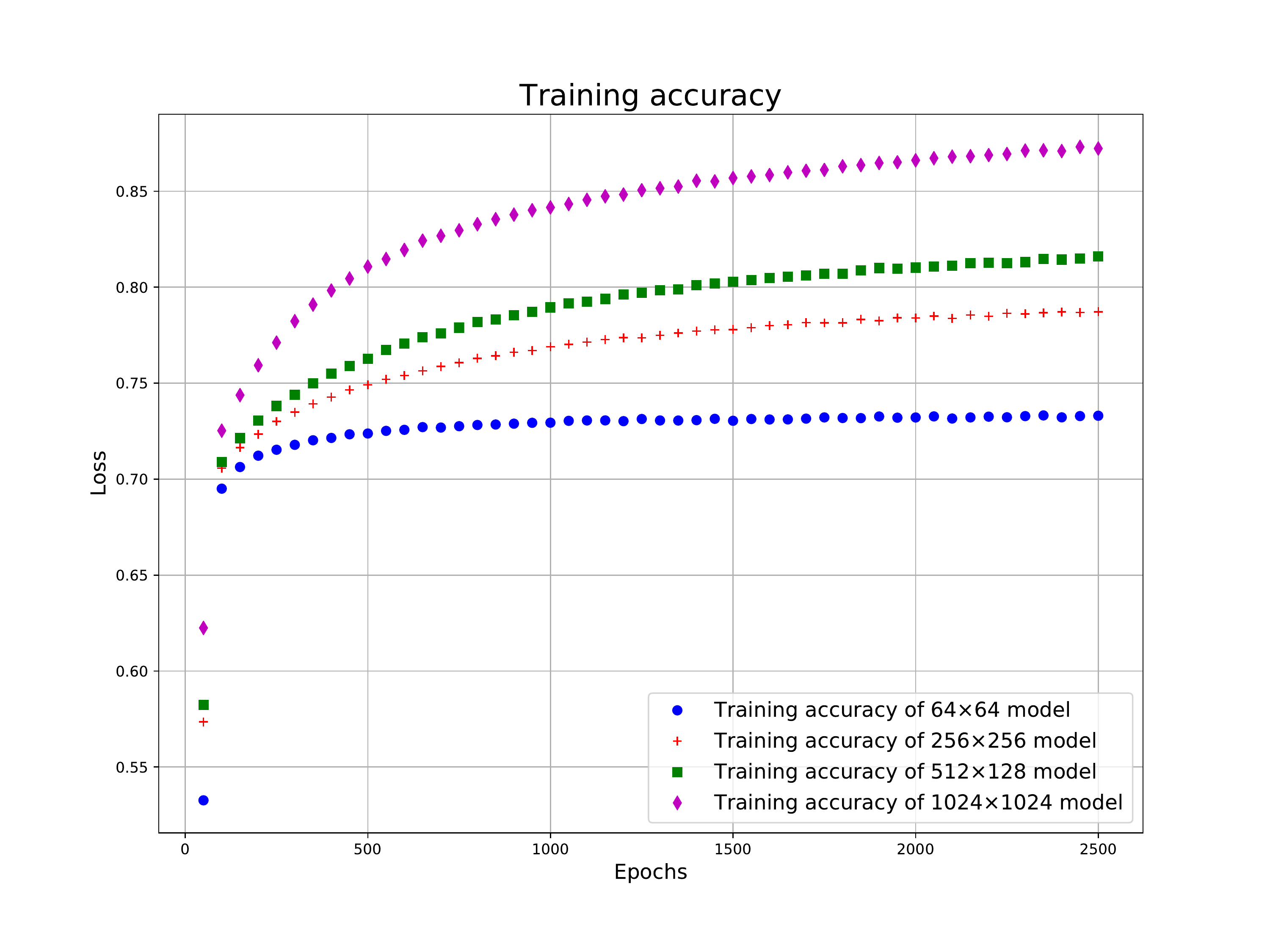}
\end{center}

\begin{center}
\begin{tabular}{ r | c  c  c  c  }
& $64 \, \times \, 64$ & $256 \, \times \, 256$ & $512 \, \times \, 128$ & $1024 \, \times \, 1024$ \\ 
\hline
\textbf{loss} & 0.5385 & 0.5194 & 0.5127 & 0.4895 \\
\textbf{accuracy} & 72.14\% & 73.74\% & 74.73\% & 76.45\% \\
\textbf{false negatives} & 13.5\% & 12.35\% & 9.32\% & 11.0\% \\
\end{tabular}
\end{center}

\section{Conclusion and discussion}
It is clear that thanks to dimensionality reduction we can create a~neural network model that can perform the~premise selection task very swiftly and with relatively high accuracy. Nevertheless, it seems that, in different applications, deep learning achieves even better results. So we could ask a~question: how the~above approach could be improved. First of all, we need to realise that the~choice of negative examples may have influenced the~performance negatively. The~machine learning algorithms generally require an~equal number of positive and negative examples, so that the~model is not biased towards predicting one more often than the~other. But as long as producing positive examples is trivial (provided that we have a~valid proof of the~conjecture), and we empirically see that the~algorithm seldom misclassifies positive examples as negatives (see the~table above), the~same cannot be said about negative examples. The~fact, that we have no proof of a~given conjecture, which would rely on some axiom, does not imply that there exist no proof depending on this axiom. Obviously, we could include, as negative examples, axioms from a~completely different theory, assuring that they are almost certainly useless. But this only weakens our model, as positive and negative examples should have similar nature, so that the~model can focus on this features which really decide whether given axiom is useful or not. So far it does not seem like there is any good solution for this dilemma.

Another problem, is the~fact that, when we focus solely on the~functional signatures of premises, we completely ignore the~logical structure of the~statements, and hence the~relations between functions. This does not happen if we use character-level representation (or even word-level representation with tokens like brackets also treated as words). But for the~neural network to clearly identify these relations, it would have to be very deep, and hence computationally inefficient - given that the~input is also highly dimensional in this setting. Another way of dealing with this issue is to substitute the~statements with graphs, where vertices represent objects, and edges the~relations between them. But this setting also requires complicated neural networks, obstructing its performance.

The~dimensionality reduction, that we adopted in this paper, is a~wonderful tool, which allows us to greatly decrease the~time required to make predictions, but it can also mean the~loss of essential information, often required to make these predictions. In natural language processing it is very likely that the~blank space in the~statement
\begin{center}
	\textsl{This is a~glass of an~orange \_\_\_\_\_.} \\
\end{center}
ought to be filled with the~word 'juice', indicating that words can often easily be deduced just from the~context, and the~loss of information, while switching from one-hot to context embeddings, is negligible. If we also have a~sentence
\begin{center}
\textsl{This is a~glass of an~apple \_\_\_\_\_.} \\
\end{center}
then it will probably be filled with the~same word. So 'orange' and 'apple' will have similar context embedding, without explicitly telling the~computer that they are fruits. Whether or not a~similar phenomenon occurs in functional signatures is debatable. Perhaps in the~future a~more natural embeddings will emerge.

And finally, having discussed the~issues with the~input data, let us deliberate on the~network architecture. Let us start with emphasising that using convolutional or recurrent networks is unsuitable in this setting. The~ordering of functions inside the~functional signature (and thus also inside the~lower dimensional embedding) is arbitrary (and we simply used the~alphabetical order), so there is no theoretical justification for the~use of convolutional neural networks, as their purpose is to identify local patterns between neighbouring objects. Also in practice, their performance appears to be inferior to fully connected networks for this task, when trained and tested on the~same data. Temporal architectures are unsuitable for the~similar reason, i.e. there is no clear temporal ordering of the~functional signatures. It is possible to slightly improve the~performance of the~model however, by including more hidden - densely connected layers. But this, while decreasing the~training time and increasing the~accuracy, also increases overfitting and the~prediction time and, making their introduction counterproductive.

\section{Acknowledgement}
The~authors of this article would like to thank the~UK Engineering and Physical Sciences Research Council (EPSRC) and the~School of Computer Science at the~University of Manchester for their financial support.

\end{document}